\documentclass[letterpaper, 10 pt, conference]{ieeeconf}  

\IEEEoverridecommandlockouts         
\let\labelindent\relax
\usepackage{enumitem}
\usepackage{amsthm}
\usepackage{color}
\usepackage{comment}
\usepackage{cite}
\usepackage[ruled,linesnumbered,vlined]{algorithm2e}

\usepackage{todonotes}
\usepackage{multicol}
\usepackage{amsmath,amssymb}

\newcommand{\E}{\mathcal{E}}

\newcommand{\real}{\mathbb{R}}

\newcommand{\N}{\mathcal{N}}

\theoremstyle{plain}
\newtheorem{theorem}{Theorem}
\newtheorem{remark}{Remark}
\newtheorem{problem}{Problem}

\newtheorem{lemma}[theorem]{Lemma}

\theoremstyle{definition}
\theoremstyle{remark}

\usetikzlibrary{decorations.pathreplacing}
\usetikzlibrary {arrows.meta}

\makeatletter
\let\NAT@parse\undefined
\makeatother
\usepackage[colorlinks = True, linkcolor = blue, citecolor = blue]{hyperref}

\begin{document}

\title{ \LARGE \bf Dynamics-aware Diffusion Models for Planning and Control}
\author{Darshan Gadginmath \quad \quad Fabio Pasqualetti \thanks{Darshan Gadginmath \href{mailto:dgadg001@ucr.edu}{(\texttt{dgadg001@ucr.edu})} and Fabio Pasqualetti \href{mailto:fabiopas@ucr.edu}{\texttt{(fabiopas@ucr.edu)}} are with the Department of Mechanical
                                Engineering, University of California Riverside, Riverside, CA, 92521, USA. 
                                }
                                }
\maketitle

\begin{abstract} 
    This paper addresses the problem of generating dynamically admissible trajectories for control tasks using diffusion models, particularly in scenarios where the environment is complex and system dynamics are crucial for practical application. We propose a novel framework that integrates system dynamics directly into the diffusion model's denoising process through a sequential prediction and projection mechanism. This mechanism, aligned with the diffusion model's noising schedule, ensures generated trajectories are both consistent with expert demonstrations and adhere to underlying physical constraints. Notably, our approach can generate maximum likelihood trajectories and accurately recover trajectories generated by linear feedback controllers, even when explicit dynamics knowledge is unavailable. We validate the effectiveness of our method through experiments on standard control tasks and a complex non-convex optimal control problem involving waypoint tracking and collision avoidance, demonstrating its potential for efficient trajectory generation in practical applications. Our code repository is available at \url{www.github.com/darshangm/dynamics-aware-diffusion}.
\end{abstract}

\section{Introduction} \label{sec:intro}
Diffusion models have emerged as powerful tools for learning complex data distributions, demonstrating significant potential in control and robotics, particularly for high-dimensional trajectory generation~\cite{TU-JL-KT:2024}. Their ability to learn and replicate expert demonstrations makes them attractive for imitation learning and decision-making. However, a critical limitation arises from their inherent lack of explicit dynamics awareness. Standard diffusion models, trained on diverse datasets, often produce trajectories that violate the underlying physical constraints of specific systems. This issue is exacerbated in robotics, where datasets often include demonstrations on different robots with varying dynamics, hindering the model's ability to generalize to individual robot behaviors. Consequently, generated trajectories may necessitate computationally expensive post-processing or real-time corrections to ensure feasibility, particularly in safety-critical applications where constraint violations can lead to failures.

To mitigate this challenge, we introduce a novel dynamics-aware diffusion framework that integrates system dynamics directly into the denoising process. By employing a sequential prediction and projection mechanism, our method ensures that generated trajectories adhere to the physical laws governing a specific robot. This approach is particularly advantageous when dealing with diverse robotic datasets, as it enables the diffusion model to specialize its output to the dynamics of a target robot. Our framework eliminates the need for external corrective mechanisms, facilitating the generation of physically plausible trajectories that are consistent with expert demonstrations. Furthermore, our approach is designed to be adaptable, offering robustness even when explicit system dynamics are partially known or approximated. By enforcing dynamics awareness, we enhance the practicality and safety of diffusion-based trajectory generation, paving the way for more reliable and efficient robotic control.

\subsection{Related work}
Diffusion models have demonstrated significant success across various domains, including image synthesis, natural language processing, and control. Their ability to model complex data distributions is particularly relevant for trajectory generation in robotics, reinforcement learning, and control. We review relevant literature in three key areas: (1) physics-informed diffusion models, (2) diffusion models for motion planning, and (3) diffusion models for control.

\noindent\textbf{Physics-informed diffusion models.} Integrating physical constraints into deep generative models is a growing research area. Many approaches incorporate physical constraints into the training or inference process by penalizing dynamics violations in the loss function~\cite{JHB-WCS-DMK:2024, DS-ZL-ABF:2023}. These methods, while effective, do not guarantee strict adherence to system dynamics. Some methods, such as~\cite{SH-etal:2024}, modify the score function to enforce conservation laws. However, these methods rely on approximate penalties rather than exact constraints. 

\noindent\textbf{Diffusion models for robotic motion planning.} Diffusion models have been applied to motion planning, particularly in imitation and reinforcement learning settings~\cite{MJ-YD-JT-SL:2022, CC-etal:2023, ZX-etal:2023, WL-XW-BJ-HZ:2023, HH-etal:2023_multitaskRL, SH-SB-KRZ-GSS:2023}. Diffuser~\cite{MJ-YD-JT-SL:2022} models multi-step trajectories as a generative process but does not explicitly incorporate system dynamics, often requiring additional controllers for correction. Similarly,~\cite{ZD-JH-etal:2024} uses iterative refinement for robotic manipulation. While~\cite{AL-ZD-ABD-RB:2024} penalizes general constraint violations in the loss function, it still does not enforce hard dynamics constraints.

\noindent\textbf{Diffusion models for control.} Diffusion models have also been explored for control tasks. For example,~\cite{GZ-etal:2024_diffusionMPC} uses a diffusion model for Model Predictive Control, and~\cite{SZ-IK-PM:2025} predicts stochastic aspects of a dynamical system for trajectory optimization. Our previous work~\cite{KE-DG-FP:2024} directly applies noising-denoising to control channels, but successfully applied to only driftless systems. The most related works are~\cite{RR-AvR-APS:2024} and~\cite{JBB-KR-etal:2025}, which use projection onto a feasible set during denoising. However, they require full knowledge of system dynamics and online computation of the feasible set at each time step. Although~\cite{{JBB-KR-etal:2025}} doesn't explicitly use system dynamics, they assume access to a perfect simulator that mimics the dynamics of the system. Furthermore, the projection step in~\cite{RR-AvR-APS:2024} and~\cite{JBB-KR-etal:2025} involves solving a nonlinear optimization problem, leading to high computational costs during inference. While~\cite{RR-AvR-APS:2024} applies projections at every denoising iteration,~\cite{JBB-KR-etal:2025} uses stochastic projections only at low noise levels. In contrast, our method can generate admissible trajectories even with unknown system dynamics. We employ a sequence of small projections through a simple matrix multiplication that is aligned with the forward diffusion noise schedule. This allows us to refine the generated trajectory while maintaining consistency with the system dynamics. 

\subsection{Contributions}
The primary contributions of our work are two-fold:
\begin{enumerate}
\item \emph{Dynamics-aware denoising via projection:} We propose a novel algorithm that integrates system dynamics into diffusion models by incorporating a projection step within the denoising process. This projection step enforces maximum-likelihood trajectory consistency with the system dynamics, augmenting the existing neural network-based denoising process. 
\item \emph{Applicability to known and unknown systems:} We demonstrate the efficacy of our approach in solving complex control problems for linear systems, showcasing adaptability across scenarios with both known and unknown system dynamics. Furthermore, we theoretically show the method's ability to recover trajectories generated by linear feedback controllers, highlighting its practical relevance.
\end{enumerate}
\section{Problem formulation and preliminaries}
\label{sec:bgrnd}
Consider the stochastic discrete-time LTI system:
\begin{equation}
    x(t+1) = Ax(t) + Bu(t) + w(t),  \label{eqn:lin_sys}
\end{equation}
where $x(t) \in \mathbb{R}^n$ represents the state vector, $u(t) \in \mathbb{R}^m$ the control input vector, and $w(t) \in \mathbb{R}^n$ is the process noise vector with mean $\mathbb{E}[w(t)] = 0$, at time $t$. The system dynamics are defined by matrices $A \in \mathbb{R}^{n \times n}$ and $B \in \mathbb{R}^{n \times m}$. The objective is to solve the following control problem:
\begin{equation}
\begin{aligned}
    \underset{u(0:T-1)}{\text{max}} \hspace*{0.25cm} & \mathcal{R}(x(0:T), u(0:T-1), \E) \\
    \text{s.t.} \quad x(t+1) &= Ax(t) + Bu(t) + w(t), \\
    x(0) &= x_{\text{init}}, \\
\end{aligned} \label{eqn:control_task}
\end{equation}
Here, $x(0:T) = [x(0)^\top \ x(1)^\top \ \dots \ x(1)^\top]^\top$ and $u(0:T-1) = [u(0)^\top \ u(1)^\top \ \dots \ u(T-1)^\top]^\top$ denote the concatenated state and control input sequences, respectively. The environmental variable $\mathcal{E}$ encapsulates task-specific information. For instance, $\E$ can encode the reward function parameters, obstacle configurations, or reference trajectories. Note that the initial state $x_0$ is fixed. We seek to determine the optimal state and control sequence $(x(0:T),u(0:T-1))$ that maximizes the expected reward $\mathcal{R}$ while adhering to the system dynamics. This problem is challenging due to potential non-convexities and randomness in the reward function $\mathcal{R}$ introduced by the environmental variable $\mathcal{E}$. Traditional control methods relying on convexity and Gaussian assumptions may be inadequate in such scenarios. 

We denote the trajectory of states and control inputs as $\tau = [x(0:T)^\top \ u(0:T-1)^\top]^\top$. To address this control challenge, we assume access to state and control trajectories of expert demonstrations, denoted as $\tau_0$. Each $\tau_0$ represents successful a task execution under different environmental conditions $\mathcal{E}$. These demonstrations are considered samples drawn from the distribution $\mathbb{P}(\mathcal{T}|\mathcal{E})$, where $\mathcal{T}$ represents the space of possible trajectories. Leveraging the generative capabilities of diffusion models, we aim to synthesize new trajectories $\tau' \sim \mathbb{P}(\mathcal{T}|\mathcal{E})$. These synthesized trajectories included control actions that can be applied to the system~\eqref{eqn:lin_sys} as they are admissible. Standard diffusion models can be conditioned with information such as the initial $x_\text{init}$, or final states $x_\text{target}$, or the environmental variables $\E$ to generate new $\tau'$, which is the reason for their effectiveness. However, diffusion models, trained solely on expert data, overlook the underlying system dynamics. Further, they're trained on diverse systems where the dynamics vary.  Consequently, generated trajectories may violate the constraints imposed by the system dynamics, rendering them physically unrealizable. In this paper, we propose a novel framework to develop dynamics-aware diffusion models by explicitly integrating the system dynamics into the generative process. We begin with a review of diffusion models.

\subsection{Denoising Diffusion Probabilistic Models (DDPMs)}
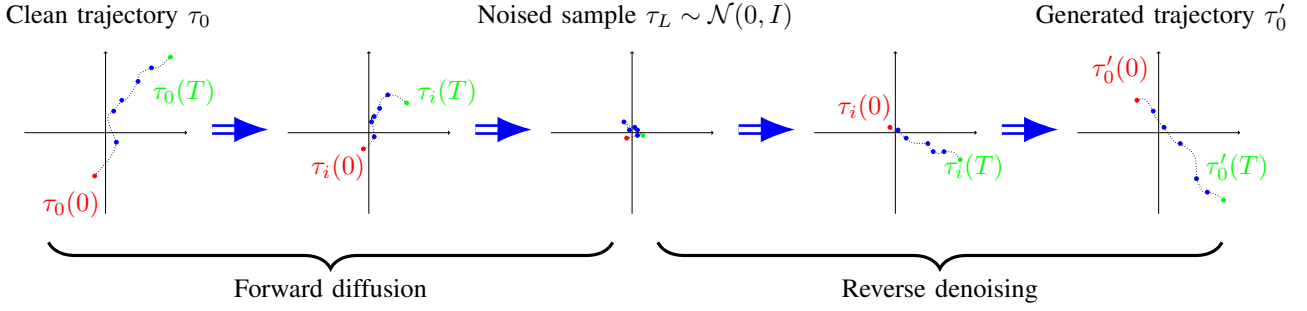
\begin{figure*}
\centering
\begin{tikzpicture}
    \node(img1) at (-7,0){\includegraphics[width=2.5cm]{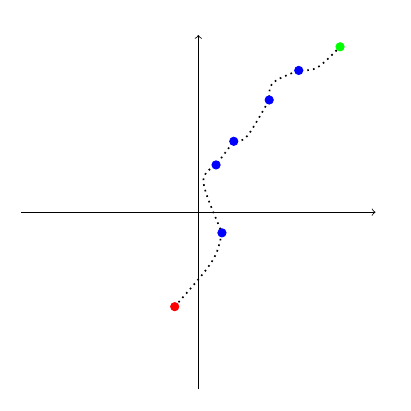}};
    \node (img2) at (-3.5,0) {\includegraphics[width=2.5cm]{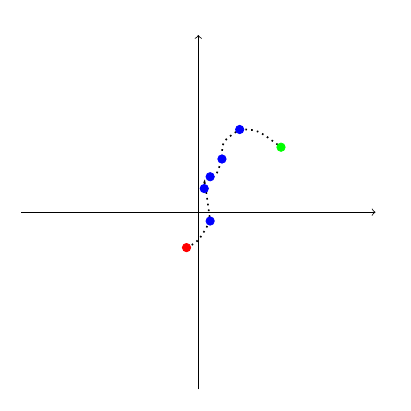}};
    \node (img3) at (0,0) {\includegraphics[width=2.5cm]{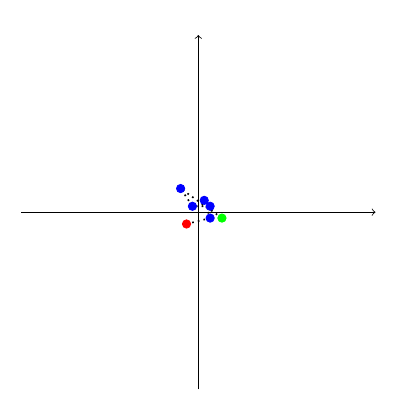}};
    \node (img4) at (3.5,0) {\includegraphics[width=2.5cm]{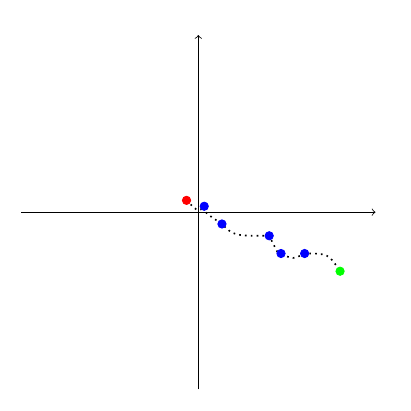}};
    \node (img5) at (7,0) {\includegraphics[width=2.5cm]{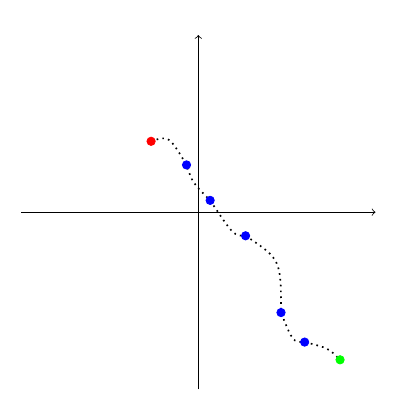}};

    \node at (-7,1.5) {Clean trajectory $\tau_0$};

    \node at (0,1.5) {Noised sample $\tau_L \sim \N(0,I)$};

    \node at (7,1.5) {Generated trajectory $\tau'_0$};
    
    \draw[line width=1pt, double distance=2pt,
    arrows = {-Latex[length=1pt 3 0]},blue] (img1) -- (img2);
    \draw[line width=1pt, double distance=2pt,
    arrows = {-Latex[length=1pt 3 0]},blue] (img2) -- (img3);
    \draw[line width=1pt, double distance=2pt,
    arrows = {-Latex[length=1pt 3 0]},blue] (img3) -- (img4);
    \draw[line width=1pt, double distance=2pt,
    arrows = {-Latex[length=1pt 3 0]},blue] (img4) -- (img5);

    \node at (-7.5,-1.0) {\textcolor{red}{$\tau_0(0)$}};
    \node at (-6.0,0.5) {\textcolor{green}{$\tau_0(T)$}};

    \node at (-3.95,-0.5) {\textcolor{red}{$\tau_i(0)$}};
    \node at (-2.5,0.5) {\textcolor{green}{$\tau_i(T)$}};

    \node at (3.05,0.3) {\textcolor{red}{$\tau_i(0)$}};
    \node at (4.5,-0.5) {\textcolor{green}{$\tau_i(T)$}};

    \node at (6.45,0.8) {\textcolor{red}{$\tau'_0(0)$}};
    \node at (8,-0.5) {\textcolor{green}{$\tau'_0(T)$}};    

    \draw[decorate, decoration={brace, amplitude=10pt, mirror}, very thick] (-7.8,-1.5) -- (-0.3,-1.5) node[midway, below=10pt] { Forward diffusion};
    \draw[decorate, decoration={brace, amplitude=10pt, mirror}, very thick] (0.3,-1.5) -- (7.8,-1.5) node[midway, below=10pt] { Reverse denoising};

\end{tikzpicture}
\caption{Example of process of forward diffusion and reverse denoising to generate new trajectories. The red, green and blue points represent the initial, final and intermediate states of a trajectory $\tau_0$, respectively. The forward diffusion process progressively adds noise to the trajectory $\tau_0$ over $L$ steps, resulting in a noisy trajectory $\tau_L$. The reverse denoising process starts from a sample of $\mathcal{N}(0,I)$ that is refinedto recover a new trajectory $\tau'_0$. }\label{fig:diffusion_process}
\end{figure*}
Diffusion models~\cite{JH-AJ-PA:2020} are generative models that learn a target data distribution $\mathbb{P}(\mathcal{T})$ by gradually denoising a sample drawn from a simple distribution, typically a Gaussian. It is inspired by non-equilibrium thermodynamics and involves two processes: a forward diffusion process and a reverse denoising process. We detail the two processes as follows.

The \emph{forward diffusion process} gradually adds Gaussian noise to the data distribution, transforming it into a simple, tractable distribution. Let $\bar{\tau} \sim \mathbb{P}(\mathcal{T})$ be a sample from the target distribution. In Figure~\ref{fig:diffusion_process}, we take the example of trajectory generation with diffusion models. The leftmost picture denotes a clean trajectory sampled from $\mathbb{P}(\mathcal{T})$. The intial, final, and intermediate points of the trajectory are represented in red, green, and blue, respectively. With $\tau_0 = \bar{\tau}$, the forward process defines a Markov chain that progressively adds Gaussian noise over $L$ steps:
\begin{align*}
    q(\tau_{1:L}|\tau_0) &= \prod_{t=1}^L q(\tau_i|\tau_{i-1}),\\
    q(\tau_i|\tau_{i-1}) &= \mathcal{N}(\tau_i; \sqrt{1 - \beta_i} \tau_{i-1}, \beta_i I).
\end{align*}
Here, $\beta_i$ is a variance schedule that controls the amount of noise added at each step $i$. The variance schedule is chosen with the following three conditions:
\begin{enumerate}[label=\textbf{(C\arabic*)}]
    \item \label{condition:init} Initial noise scale $\beta_0= 0$, and final noise scale $\beta_L = 1$. 
    \item Bounded noise scale, i.e., $\beta_i \in [0, 1]$ for all $i$.\label{condition:bound}
    \item Monotonically increasing noise scale, i.e., $\beta_i \leq \beta_{i+1}$ for all $i$.\label{condition:increasing}

\end{enumerate}
The above three conditions for the variance schedule is satisfied by simple linear schedules such as $\beta_i = k~i$, where $k$ is a constant. This variance schedule ensures that $\tau_L$ is a sample from the standard Gaussian. We can also view the forward process through the lens of dynamical systems as,
\begin{align*}
\tau_i = \sqrt{1 - \beta_i} \tau_{i-1} + \sqrt{\beta_i} \epsilon_i, \quad \epsilon_i \sim \mathcal{N}(0, I).
\end{align*}

Continuing the example of trajectory generation, we illustrate the forward diffusion process in Figure~\ref{fig:diffusion_process} on the left half. We can see that the forward diffusion process transforms the original trajectory $\bar{\tau}$ over $L$ steps, where the noise progressively obscures the original data. After the final noising step $L$, every point of the trajectory is just a sample from the standard Gaussian distribution.

The \emph{reverse denoising process} aims to learn the reverse Markov chain $p_{\theta}(\tau_{0:L})$ using a neural network parameterized by $\theta$. The parameterized model gradually removes noise from a sample $\tau_L \sim \mathcal{N}(0,I)$ to recover $\tau'_0 \sim \mathbb{P}(\mathcal{T})$ as follows:
\begin{align*}
    p_{\theta}(\tau_{0:L}) &= p(\tau_L) \prod_{i=1}^L p_{\theta}(\tau_{i-1}|\tau_i),\\
    p_{\theta}(\tau_{i-1}|\tau_i) &= \mathcal{N}(\tau_{i-1}; \mu_{\theta}(\tau_i, i), \beta_i I).
\end{align*}
The mean $\mu_{\theta}(\tau_i, i)$ is learned by the neural network, whereas the variance is fixed to $\beta_i$. We can express this as:
\begin{align}
\tau_{i-1} = \mu_{\theta}(\tau_i, i) + \sqrt{\beta_i} \epsilon_i, \quad \epsilon_i \sim \mathcal{N}(0, I). \label{eqn:reverse_process}
\end{align}

The neural network $\mu_{\theta}(\tau_i, i)$ is trained to minimize the following loss function:
\begin{equation}
    \mathcal{L}(\theta) = \mathbb{E}_{i, \tau_0, \epsilon} \left[ \left\| \mu_{q}(\tau_{i-1}|\tau_i, \tau_0) - \mu_{\theta}(\tau_i, i) \right\|^2 \right].\label{eqn:loss_fn_diffusion}
\end{equation}
Here, $\mu_{q}(\tau_{i-1}|\tau_i, \tau_0)$ is the conditional mean of the reverse process given by:
\begin{equation*}
\mu_{q}(\tau_{i-1}|\tau_i, \tau_0) = \frac{1}{\sqrt{\alpha_i}}\left(\tau_i - \frac{\beta_i( \tau_i - \sqrt{\bar{\alpha}_i} \tau_0 )}{1-\bar{\alpha}_i}\right),
\end{equation*}
where $\alpha_i = 1 - \beta_i$ and $\bar{\alpha}_i = \prod_{s=1}^i \alpha_s$. 
Once trained, new samples $\tau'_0$ can be generated by sampling $\tau_L \sim \mathcal{N}(0, I)$ and iteratively denoising using the learned reverse process. In Figure~\ref{fig:diffusion_process}, the right half illustrates the reverse denoising process, where the noisy trajectory $\tau_L$ is progressively refined to recover a new trajectory $\tau'_0$.

\subsection{DDPMs for control}
We now formally state the problem of trajectory generation to solve~\eqref{eqn:control_task}. We have access to a dataset composed of expert demonstrations $\tau_0$ that achieve the control task~\eqref{eqn:control_task} with corresponding conditions $\E$ and $x_\text{init}$. However, we do not have access to the structure of the reward function $\mathcal{R}(x(0:T),u(0:T-1),\E)$. The control problem using the DDPM-based approach is as follows:

\begin{problem}[\textbf{Admissible trajectory generation with known dynamics}]
Given the matrices $(A,B)$, without the knowledge of the reward function $\mathcal{R}$, generate a new trajectory $\tau'$ that satisfies the system dynamics~\eqref{eqn:lin_sys} and solve the control problem in \eqref{eqn:control_task} for a given condition $\mathcal{E}$, and $x_\text{init}$. 
\end{problem}

Our second problem corresponds to the case when the system matrices $(A,B)$ are unknown. In this case, we assume we have access to one long experiment $\Gamma$ of length $S > T$, such that $\Gamma = [x(0:S)^\top| u(0:S-1)^\top]^\top$. Given the experimental trajectory $\Gamma$, the problem of generating admissible trajectories is as follows.

\begin{problem}[\textbf{Admissible trajectory generation with unknown dynamics}]
    Given only the experimental trajectory $\Gamma$, without the knowledge of the reward function $\mathcal{R}$, generate a new trajectory $\tau'$ that satisfies the system dynamics~\eqref{eqn:lin_sys} and solves the control problem in \eqref{eqn:control_task} for a given condition $\mathcal{E}$, and initial state $x_\text{init}$.
    \end{problem}



\section{Dynamics-Aware Diffusion Models}
\label{sec:method}
We describe our approach of making diffusion models dynamics-aware in this section. We first present our framework for the case when the system dynamics are known, followed by the case when the system dynamics are unknown.
\subsection{Problem 1 - Known model}
For the linear system defined in \eqref{eqn:lin_sys}, we can express the state trajectory over the horizon $[0,T]$ as
\begin{align*}
x(0:T)
&= \mathcal{A} x(0) + 
\mathcal{C}_T u(0:T-1)
+ \mathcal{C}^w_T w(0:T-1).
\end{align*}
Here, matrices $\mathcal{A}$ and $\mathcal{C}_T$ are the free and forced response matrices of the system respectively. They correspond to
\begin{align*}
    \mathcal{A} = \begin{bmatrix}
        I \\ A \\ A^2 \\ \vdots \\ A^{T}
    \end{bmatrix}, \quad
    \mathcal{C_T} = 
    \begin{bmatrix}
        0 & 0 & \cdots & 0 \\ 
        B & 0 & \cdots & 0 \\ 
        AB & B &\cdots & 0 \\ 
        \vdots & \vdots & \ddots & \vdots \\ 
        A^{T-1}B & A^{T-2}B & \cdots & B 
    \end{bmatrix} .
\end{align*}
The forced response matrix corresponding to the noise $\mathcal{C}^w_T$ is obtained by replacing $B$ with $I$ in $\mathcal{C}_T$. Using the above formulation, we can express the entire state-control trajectory $\tau = [x(0:T)^\top | u(0:T-1)^\top]^\top$ as:
\begin{align}
    \tau = \underbrace{\begin{bmatrix} \mathcal{A} & \mathcal{C}_T \\ 0 & I \end{bmatrix}}_{\mathcal{F}} \begin{bmatrix}x(0) \\\hline u(0:T-1) \end{bmatrix} +  \underbrace{\begin{bmatrix}
    \mathcal{C}^w_T \\ 0 \end{bmatrix}}_{\mathcal{F}^w}
        w(0:T-1).\label{eqn:behaviors}
\end{align}
We are interested in generating a new trajectory $\tau'$ of the system~\eqref{eqn:lin_sys} using a diffusion model such that $\tau'$ solves the control task~\eqref{eqn:control_task}. 

Let us denote a trajectory generated by the diffusion model without any dynamics awareness as $\hat{\tau}$ and the trajectory with dynamics awareness as $\tau'$. To satisfy the dynamics of the system~\eqref{eqn:lin_sys}, we now describe the projection step that we incorporate in the denoising procedure. Note that system~\eqref{eqn:lin_sys} is stochastic, so we seek to find the maximum-likelihood trajectories that are admissible. If we are given an arbitrary vector $\hat{\tau} \in \mathbb{R}^{n(T+1) + mT}$, we need to find the closest plausible trajectory of the linear system~\eqref{eqn:lin_sys}. Particularly, we need to solve the following least-squares problem:
\begin{align}
    \tau' = \underset{\tau \in S}{\text{arg~min}}~\mathbb{E}_{w_t}\|\tau - \hat{\tau}\|^2 = \mathcal{F} \mathcal{F}^\dagger \hat{\tau},\label{eqn:least_squares}
\end{align} 
where $S$ is the space of the system's trajectories and $\mathcal{F}^\dagger$ denotes the pseudo-inverse of $\mathcal{F}$. Since system~\eqref{eqn:lin_sys} is stochastic, $\tau'$ is the maximum-likelihood trajectory w.r.t $\hat{\tau}$. The solution to the least-squares problem~\eqref{eqn:least_squares} is obtained by projecting $\hat{\tau}$ onto the image of $\mathcal{F}$. This fact arises from the zero-mean assumption of the process noise, i.e., $\mathbb{E}[w_t] = 0$. Thus, to ensure that the diffusion model incorporates the dynamics in $\tau'$, one can project $\hat{\tau}$ onto the image of $\mathcal{F}$ as $\tau' = \mathcal{F} \mathcal{F}^\dagger \hat{\tau}.$
Note that projecting $\hat{\tau}$ after the entire denoising process is finished can lead to large residuals of $\|\hat{\tau} - \tau'\|$. This could cause the trajectory to  deviate from achieving the control task. Therefore, we incorporate a sequential projection step in the reverse denoising process of the diffusion model. The key idea is to modify the reverse denoising process to include a projection step at every instance $i$ that finally ensures the generated trajectory $\tau$ remains within the image of $\mathcal{F}$.

The denoising process of our diffusion model starts by sampling $\tau_L$ from a standard Gaussian distribution $\mathcal{N}(0, I)$. To emphasize that our diffusion model is conditioned on the initial state $x_\text{init}$ and the environmental variable $\E$, we use the notation $\mu_\theta(\tau'_i, i,x_\text{init},\E)$. At any denoising step $i<L$, the trajectory $\tau'_i$ is first denoised using the learned mean $\mu_{\theta}(\tau'_i, i,x_\text{init},\E)$, and then projected onto the image of $\mathcal{F}$ as,
\begin{align}
\begin{split}
    \hat{\tau}_{i-1} &= \mu_\theta(\tau'_i, i,x_\text{init},\E) + \sqrt{\beta_i} \epsilon_i, \\
    \tau'_{i-1} &= \left(\sqrt{1 - \beta_{i-1}} \mathcal{F} \mathcal{F}^\dagger  + \sqrt{\beta_{i-1}} I \right) \hat{\tau}_{i-1}.
\end{split}\label{eqn:dyn_aware_denoising}
\end{align}
We initially have a prediction from the neural network followed by a scaled projection onto the image of $\mathcal{F}$. Note that the projection is scaled by $\sqrt{1-\bar{\alpha}_i}$. This scaling balances the projection onto the admissible trajectory space with the noisy trajectory $\hat{\tau}_i-1$, ensuring a gradual convergence to the system's dynamics. Our choice of scaling is motivated by the conditions~\ref{condition:init}-\ref{condition:increasing}, where which can be used to express a time-varying dynamical system for denoising. Particularly, from conditions \ref{condition:init}-\ref{condition:increasing}, we have $1-\bar{\alpha}_i \rightarrow 1$ as $i \rightarrow 0$. This process of first denoising and then projecting ensures that the generated trajectory $\tau'_i$ converges to the space of admissible trajectories at each denoising step. The final generated trajectory $\tau'_0$ is obtained by iterating this process for $L$ steps.  We elucidate our method in Algorithm~\ref{algo:denoising-known-dynamics}. In the following result, we show that the generated trajectories $\tau'_0$ of the proposed method satisfy the system dynamics in \eqref{eqn:lin_sys}.
\begin{algorithm}[t]
    \label{algo:denoising-known-dynamics}
    \caption{Generating admissible trajectories with known dynamics}
    \KwData{ $x_\text{init}$, $\mathcal{E}$, $\mathcal{F}$}
    \SetKwInOut{Input}{Initialize}
    \SetKwInOut{Output}{Output}
    \Input{$L$ denoising steps, noising schedule $\beta_i$ for $i=1,\dots,L$, neural network $\mu_\theta(\tau_i,i,x_\text{init},\E)$}
    \SetAlgoLined
    Sample $\tau'_L \sim \mathcal{N}(0,I)$ \\
    \For{$i \gets L$ \KwTo $1$}{
    Predict: $\hat{\tau}_{i-1} = \mu_\theta(\tau'_i, i,x_\text{init},\E) + \sqrt{\beta_i} \epsilon_i$ \\
    Project: $\tau'_{i-1} = \left(\sqrt{1 - \beta_{i-1}} \mathcal{F} \mathcal{F}^\dagger  + \sqrt{\beta_{i-1}} I \right) \hat{\tau}_{i-1} $
    }
    \Output{Trajectory $\tau'_0$.}
\end{algorithm}

\begin{lemma}[\textbf{Generation of admissible trajectories}]
    Given the matrices $(A,B)$ for system~\eqref{eqn:lin_sys}, the generated trajectory $\tau'_0$ using Algorithm~\ref{algo:denoising-known-dynamics} is the maximum likelihood trajectory. 
\end{lemma}
\noindent\textbf{Proof.}
We highlight the key steps of the proof. Conditions \ref{condition:init}-\ref{condition:increasing} ensure that $\beta_i \rightarrow 0$ as $i \rightarrow 0$. This ensures that trajectory $\tau'_i$ gets closer to the image of $\mathcal{F}$. At the final denoising step $i=1$ of Algorithm~\ref{algo:denoising-known-dynamics}, we have $\beta_0= 0$. The final generation step is given by:
\begin{align}
    \begin{split}
    \hat{\tau}_0 &= \mu_\theta(\tau'_1,1,x_\text{init},\E) + \sqrt{\beta_1} \epsilon_1, \\
    \tau'_0 &= \mathcal{F} \mathcal{F}^\dagger \hat{\tau}_0. 
    \end{split} \label{eqn:final_denoising}
\end{align}
 This implies that $\tau'_0$, as shown in~\eqref{eqn:least_squares}, is the maximum-likelihood trajectory after the final denoising step. \hfill $\blacksquare$

The above lemma shows that the generated trajectory $\tau'_0$ using Algorithm~\ref{algo:denoising-known-dynamics} is admissible and satisfies the system dynamics in \eqref{eqn:lin_sys}. For a special case where the system is noiseless, $\mathcal{F}^w = 0$, the generated trajectory $\tau'_0$ exactly follows the system dynamics. We still need to ensure that the generated trajectory $\tau'_0$ also solves the control problem in \eqref{eqn:control_task}. To achieve this, we need Algorithm~\ref{algo:denoising-known-dynamics} to generate trajectories from $\mathbb{P}(\mathcal{T}|E)$. We show that Algorithm~\ref{algo:denoising-known-dynamics} achieves this by minimizing a measure of the distance between $\tau'_i$ and $\mathbb{P}(\mathcal{T}|\E)$. In the following result, we show that Algorithm~\ref{algo:denoising-known-dynamics} generates a trajectory $\tau'_0$ by sequentially minimizing the Mahalanobis distance to the target density $\mathbb{P}(\mathcal{T}\E)$.

\begin{theorem}[\textbf{Generation of trajectories for linear feedback control}] \label{thm:denoising_linear_feedback_control}
    Consider an LTI system as in~\eqref{eqn:lin_sys} with white Gaussian process noise $w_t \sim \mathcal{N}(0,I)$, where the optimal solution to control problem~\eqref{eqn:control_task} is a linear feedback controller $u(t) = K(t|\E) x(t)$. Algorithm~\ref{algo:denoising-known-dynamics} generates trajectories that correctly sample from the distribution of optimal trajectories of the system $\mathbb{P}(\mathcal{T}|\E)$ by sequentially minimizing the Mahalanobis distance:
    \begin{align}
        d(\tau'_i|\mathbb{P}(\mathcal{T}|\E)) = \left\| \tau'_i - \mu_{\mathcal{T}|\E} \right\|^2_{\Sigma_{\mathcal{T}|\E}^{-1}}
    \end{align}
    where $\mu_{\mathcal{T}|\E}$ and $\Sigma_{\mathcal{T}|\E}$ represent the mean and covariance of the distribution of optimal trajectories $\mathbb{P}(\mathcal{T}|\E)$, such that:
    \begin{align}
        d(\tau'_{i-1}|\mathbb{P}(\mathcal{T}|\E)) \leq d(\tau'_i|\mathbb{P}(\mathcal{T}|\E)).
    \end{align}
\end{theorem}
\noindent\textbf{Proof.} 
We begin by analyzing the statistical properties of $\mathbb{P}(\mathcal{T}|\E)$. If the optimal solution to the control problem~\eqref{eqn:control_task} is a linear feedback controller $u(t) = K(t|\E)x(t)$, then any trajectory $\tau$ can be expressed as:
\begin{align}
    \tau = \begin{bmatrix}
        x(0) \\ x(1) \\ \vdots \\ x(T) \\ \hline K(0|\E)x(0) \\ K(1|\E)x(1) \\ \vdots \\ K(T-1|\E)x(T-1)
    \end{bmatrix}
\end{align}
For brevity, let us denote $\tilde{A}(t) = A-BK(t|\E)$. Then the dynamics of the mean of the states is $\mathbb{E}[x(t+1)] = \tilde{A}(t)  \mathbb{E}[x(t)]$, and the covariance is $\mathbb{C}\text{ov}[x(t+1)] = \tilde{A}(t)  \mathbb{C}\text{ov}[x(t)]  \tilde{A}(t) ^{\top} + I$. Similarly, the control inputs have mean $\mathbb{E}[u(t)] = K(t|\E)\mathbb{E}[x(t)]$, and covariance $\mathbb{C}\text{ov}[u(t)] = K(t|\E)\mathbb{E}[x(t)x(t)^\top] K(t|\E).$
Thus, we can express the distribution of optimal trajectories $\mathbb{P}(\mathcal{T}|\E)$ as a Gaussian distribution with mean $\mu_\mathcal{T|\E}$ and covariance $\Sigma_\mathcal{T|\E}$, where the mean and covariance are obtained from the statistics of the state and control inputs. Since, $\mathbb{P}(\mathcal{T}|\E)$ is Gaussian, minimizing the Mahalanobis distance $d(\tau'_i|\mathbb{P}(\mathcal{T}|\E))$ is equivalent to maximizing the likelihood of the generated trajectory $\tau'_i$ with respect to the distribution of optimal trajectories $\mathbb{P}(\mathcal{T}|\E)$.

Now, examining Algorithm~\ref{algo:denoising-known-dynamics}, we observe that each denoising step performs a projection onto the space of valid system dynamics. At each denoising step $i$, the algorithm predicts a noisy sample $\tau'_{i-1}$ using the neural network $\mu_\theta(\tau'_{i},i,x_\text{init},\E)$. Then it projects the sample $\tau'_{i-1}$ onto the space of valid trajectories by enforcing the linear dynamics.

For the prediction step, the neural network $\mu_\theta(\tau'_i, i,x_\text{init},\E)$ is trained to approximate the mean of the forward process $q(\tau_{i-1}|\tau_i, \tau_0)$. This prediction is based on the current state $\tau'_i$ and the initial state $x_\text{init}$, along with any environmental variables $\E$. The second step involves projecting the predicted trajectory onto the space of valid trajectories defined by the system dynamics using the scaled projection given in~\eqref{eqn:dyn_aware_denoising}. This projection is equivalent to finding the trajectory that minimizes the Mahalanobis distance to the distribution $\mathbb{P}(\mathcal{T}|\E)$. To see this, note that the algorithm's projection step can be formulated as:
\begin{align*}
    \tau'_{i-1} = \underset{\tau \in \mathcal{S}}{\arg\min} \|\tau - \hat{\tau}_{i-1}\|^2
\end{align*}
where $\mathcal{S}$ is the space of admissible trajectories of the system. Since $\mathbb{P}(\mathcal{T}|\E)$ is Gaussian, this projection precisely minimizes the Mahalanobis distance $d(\tau'_i|\mathbb{P}(\mathcal{T}|\E)) = \|\tau'_i - \mu_{\mathcal{T}|\E}\|^2_{\Sigma_{\mathcal{T}|\E}^{-1}}$. As the diffusion process proceeds from $i=L$ to $i=0$, $\beta_i$ decreases to zeros. Hence, the sequential application of these projections ensures that the final trajectory $\tau'_0$ correctly samples from the distribution of optimal trajectories. Each projection operation brings the trajectory closer to the space of valid system trajectories. Specifically, at each step $i$, the Mahalanobis distance decreases as:
\begin{align*}
    d(\tau'_{i-1}|\mathbb{P}(\mathcal{T}|\E)) &= \|\tau'_{i-1} - \mu_{\mathcal{T}|\E}\|^2_{\Sigma_{\mathcal{T}|\E}} \\
    &= \Big|\Big|\left(\sqrt{1 - \beta_{i-1}} \mathcal{F} \mathcal{F}^\dagger + \sqrt{\beta_{i-1}} I \right) \hat{\tau}_{i-1} \\ 
    & \hspace*{0.38cm}  - \mu_{\mathcal{T}|\E}\Big|\Big|^2_{\Sigma_{\mathcal{T}|\E}^{-1}} \\
    &\leq \|\hat{\tau}_{i-1} - \mu_{\mathcal{T}|\E}\|^2_{\Sigma_{\mathcal{T}|\E}} \\
    &\leq \|\tau'_i - \mu_{\mathcal{T}|\E}\|^2_{\Sigma_{\mathcal{T}|\E}} = d(\tau'_i|\mathbb{P}(\mathcal{T}|\E))
\end{align*}
The first inequality follows from the optimality of the projection operator. The second inequality follows from the fact that the neural network is trained to approximate the mean of the posterior distribution~\eqref{eqn:loss_fn_diffusion}, and the projection operation reduces the distance to the space of valid trajectories.  \hfill $\blacksquare$

Theorem~\ref{thm:denoising_linear_feedback_control} highlights the method's applicability to a wide range of control problems where the optimal solution is a linear state feedback. This includes classic problems like the finite or infinite horizon Linear Quadratic Regulator (LQR). The theorem demonstrates that our approach effectively generates trajectories for such problems by sequentially minimizing the Mahalanobis distance between the generated trajectory and the distribution of optimal solutions. Theorem~\ref{thm:denoising_linear_feedback_control} can be generalized to any control problem where the optimal solution is an affine feedback of the form $u(t) = K(t|\E) x(t) + c(t)$, which corresponds to tracking problems.

\subsection{Problem 2 - Unknown model}
We now present our algorithm when we do not have access to the matrices $(A,B)$. In the absence of known system matrices, we utilize the data-driven approach enabled by Willems' Fundamental Lemma, effectively replacing the system's free and forced response matrices with Hankel matrices constructed from experimental data. We assume access to a long experiment $\Gamma$ of length $S>T$, such that $\Gamma = [x(0:S)^\top | u(0:S)^\top]^\top$. We begin by recalling the notions of data-driven control of linear systems. A control signal $u(0:S)$ is said to be persistently exciting of the order $T$ if the Hankel matrix $H_T(u)$ defined as 
\begin{align}
    H_T(u) = \begin{bmatrix}
        u(0) & u(1) & \cdots & u(S-T+1) \\
        u(1) & u(2) & \cdots & u(S-T+2) \\
        \vdots & \vdots & \ddots & \vdots \\
        u(T-1) & u(T) & \cdots & u(S),
    \end{bmatrix}
\end{align}
has full row rank. Persistently exciting control inputs ensures that the columns of the Hankel matrix constructed resulting state trajectory $H_T(x)$ spans the space of possible states, allowing for accurate reconstruction of system trajectories. From Willem's Fundamental Lemma~\cite{JCW-PR-IV-BLMdM:2005}, we can construct new trajectories of the system using Hankel matrices of the states and control inputs. In particular, we can express any trajectory $\tau$ of the system as:
\begin{align}
    \tau = \begin{bmatrix}
        H_{T+1}(x) \\ H_T(u)
    \end{bmatrix}g,
\end{align}
where $g \in R^{n(T+1) + m(T)}$, is an arbitrary vector and the Hankel matrices $H_{T+1}(x)$ and $H_T(u)$ are constructed from the long experiment $\Gamma$. For the case with process noise $w_t$, such that $\mathbb{E}[w_t] = 0$, $\tau'$ represents the maximum-likelihood trajectory corresponding to the vector $g$. If the diffusion model provides an arbitrary vector $\hat{\tau}$ without dynamics awareness, the solution to the least-squares problem~\eqref{eqn:least_squares} is
\begin{align}
    \tau' = \begin{bmatrix}
        H_{T+1}(x) \\ H_T(u)
    \end{bmatrix}
    \begin{bmatrix}
        H_{T+1}(x) \\ H_T(u)
    \end{bmatrix}^\dagger \hat{\tau}. \label{eqn:projection_unknown}
\end{align}
From here, we can proceed similar to Algorithm~~\ref{algo:denoising-known-dynamics} by modifying the projection step. The key idea is to use the projection based on Hankel-matrices as follows:
\begin{align}
    \begin{split}
        \hat{\tau}_{i-1} &= \mu_\theta(\tau'_i, i,x_\text{init},\E) + \sqrt{\beta_i} \epsilon_i, \\
        \tau'_{i-1} &= \Bigg(\sqrt{1 - \beta_{i-1}} \begin{bmatrix}
            H_{T+1}(x) \\ H_T(u)
        \end{bmatrix}
        \begin{bmatrix}
            H_{T+1}(x) \\ H_T(u)
        \end{bmatrix}^\dagger  \\& \ \ \ + \sqrt{\beta_{i-1}} I \Bigg) \hat{\tau}_{i-1}. 
    \end{split}\label{eqn:dyn_unknown_aware_denoising}
    \end{align}
This ensures that the generated trajectory $\tau'_0$ satisfies the system dynamics in a least-squares sense and solves the control problem~\eqref{eqn:control_task}. We summarize our approach in Algorithm~\ref{algo:denoising-unknown-dynamics}.
\begin{algorithm}[t]
    \label{algo:denoising-unknown-dynamics}
    \caption{Generating admissible trajectories with \emph{unknown} dynamics}
    \KwData{ $x_\text{init}$, $\mathcal{E}$, long experiment $\Gamma$}
    \SetKwInOut{Input}{Initialize}
    \SetKwInOut{Output}{Output}
    \Input{$L$ denoising steps, noising schedule $\beta_i$ for $i=1,\dots,L$, neural network $\mu_\theta(\tau_i,i)$}
    \SetAlgoLined
    Construct matrices $H_{T+1}(x)$ and $H_T(u)$ using $\Gamma$ \\
    Sample $\tau'_L \sim \mathcal{N}(0,I)$ \\
    \For{$i \gets L$ \KwTo $1$}{
    Predict: $\hat{\tau}_{i-1} = \mu_\theta(\tau'_i, i,x_\text{init},\E) + \sqrt{\beta_i} \epsilon_i$ \\
    Project with hankel matrices~\eqref{eqn:dyn_unknown_aware_denoising}:
    \begin{align*}
        \begin{split}
            \tau'_{i-1} &= \Bigg(\sqrt{1 - \beta_{i-1}} \begin{bmatrix}
                H_{T+1}(x) \\ H_T(u)
            \end{bmatrix}
            \begin{bmatrix}
                H_{T+1}(x) \\ H_T(u)
            \end{bmatrix}^\dagger  \\& \ \ \ + \sqrt{\beta_{i-1}} I \Bigg) \hat{\tau}_{i-1}. 
    \end{split}
    \end{align*}
    }
    \Output{Trajectory $\tau'_0$.}
\end{algorithm}
It is important to note that the results of Lemma 1 and Theorem 1 still hold for the case when the system dynamics are unknown. The only difference is that we now use the Hankel matrices $H_{T+1}(x)$ and $H_T(u)$ to project the predicted trajectory $\hat{\tau}$ onto the image of the system dynamics. While Willems' Fundamental Lemma is rigorously derived for noiseless linear systems, we extend its application to stochastic systems by employing a least-squares projection using the pseudo-inverse of the Hankel matrix. This approach provides a maximum-likelihood estimate of the system trajectory under the assumption of white process noise. It is important to acknowledge that noise introduces inaccuracies in trajectory reconstruction. However, longer lengths $S$ of the experimental trajectory $\Gamma$ can be used for better approximation of the system dynamics.
\begin{remark}(\textbf{Nonlinear systems and predictive control})
    Enforcement of admissible trajectories via projection for nonlinear systems typically requires solving a computationally intensive nonlinear optimization problem at each denoising step. our framework offers a computationally efficient alternative. By leveraging linearizing transformations, such as feedback linearization~\cite{DG-VK-FP:2024} or Koopman-based approximations~\cite{ZW-RMJ:2020}, we can extend our method to nonlinear systems within a model predictive control (MPC) paradigm. These transformations, accurate over short horizons, are well-suited for MPC implementations. The diffusion model receives environmental variables $\E$ as time-varying inputs and generates trajectories in the transformed linear space. Subsequent iterative sampling and inverse transformation efficiently yield trajectories in the original nonlinear coordinates, circumventing the need for computationally expensive nonlinear optimization during denoising.
\end{remark}

\section{Experimental Results}
We evaluate our proposed method on a test bed of different control problems. We compare our methods against a vanilla diffusion model with no dynamics awareness. Our experiments were conducted on an Intel i9-9900 machine with 128GB RAM and an Nvidia Quadro RTX 4000 GPU. Our code repository is available at \url{www.github.com/darshangm/dynamics-aware-diffusion}.

\subsection{LQR with 4 dimensional system}\label{sec:lqr}
In this experiment, we consider problem of generating trajectories of an LTI system for the Linear Quadratic Regulators (LQR) problem. The system~\eqref{eqn:lin_sys} has the matrices:
\begin{align}
A = \begin{bmatrix}
    1 & 0 & 0.1 & 0 \\
    0 & 1 & 0 & 0.1 \\
    0 & 0 & 1 & 0 \\
    0 & 0 & 0 & 1\end{bmatrix}, \quad B = \begin{bmatrix}
        0 & 0 \\
        0 & 0 \\
        0.1 & 0 \\
        0 & 0.1
    \end{bmatrix},
\end{align}
which corresponds to the dynamics of a double integrator discretized with a time step of 0.1$s$. The LQR problem is,
\begin{align}
    \begin{split}
        &\min_{u(t)} \quad \mathbb{E}\left[ \sum_{t=0}^{T-1} \|x(t) - x_\text{target}\|_Q^2  + \|u(t)\|_R^2 \right] \\
        &\quad \text{s.t. }\quad x(t+1) = Ax(t) + Bu(t) + w(t),\\
        & \quad \quad \quad \hspace*{0.75cm} x(0) = x_\text{init}
    \end{split}
\end{align}
The optimal controller for this problem is given by $u(t) = K(t)x(t) + c(t)$, where $K(t)$ is the feedback gain matrix and is a constant feedforward term. We generate a synthetic dataset of 10,000 trajectories of length $T =30$, where each trajectory is generated by sampling the initial state $x_\text{init}$ from a uniform distribution over $\mathcal{U}[-1,1]^4$ and $x_\text{target}$ from a uniform distribution over $\mathcal{U}[-4,4]^4$. The control input $u(t)$ is generated using the LQR controller with $R = I$ and $Q$ is a diagonal matrix such that $Q = \text{diag}[10,10,1,1]$. For each of the three algorithms (vanilla diffusion, our methods with known dynamics and unknown dynamics), we use the same neural network architecture which consists of an encoder-decoder network with 3 convolutional layers with 256 hidden units. The encoder takes the trajectory $\tau$ as input, with time and condition embeddings and outputs a latent representation of the trajectory. The decoder takes the latent representation and generates the trajectory. The neural network is trained using the Adam optimizer . The diffusion model is trained for 30,000 epochs with a batch size of 64. We use a linear noising schedule with $\beta_i = 0.001i$ for $i=1,\dots,L=1000$. The diffusion model is conditioned on the initial state $x_\text{init}$ and the environmental variable $\E$. For algorithm 3, we generate a long experiment $\Gamma$ of length $S=100$ using $u(t)\sim  \mathcal{N}(0,I)$.
\begin{figure}
    \centering
    \begin{tikzpicture}
        \node (img1) at (-3.65,0) {\includegraphics[width=0.220\textwidth]{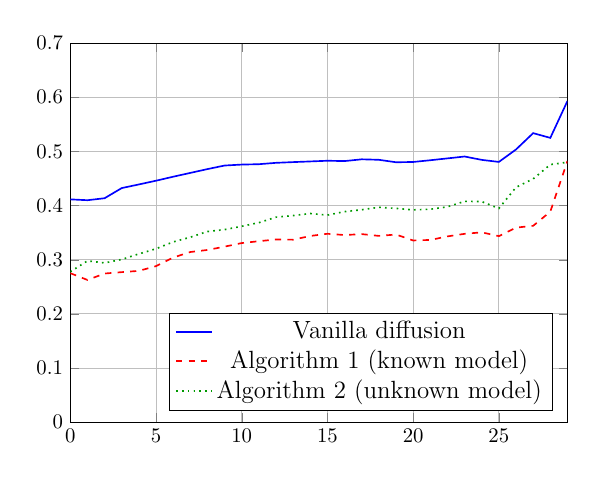}};
        \node[below of= img1, node distance=0cm, xshift = 0.0cm, yshift=-2.3cm,font=\color{black}]  {(a)};
        \node[below of= img1, node distance=0cm, xshift = 0.0cm, yshift=-1.8cm,font=\color{black}]  {time (t)};
        \node[left of= img1, node distance=0cm, xshift = -2.1cm, yshift=0, rotate = 90,font=\color{black}]  {Avg. $\|x(t) - x_\text{LQR}(t)\|^2$};
        \node (img2) at (0.6,0) {\includegraphics[width=0.220\textwidth]{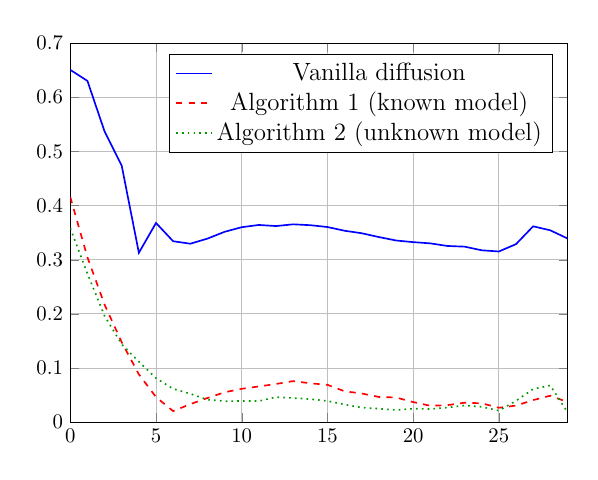}};
        \node[below of= img2, node distance=0cm, xshift = 0.0cm, yshift=-1.8cm,font=\color{black}]  {time (t)};
        \node[below of= img2, node distance=0cm, xshift = 0.0cm, yshift=-2.3cm,font=\color{black}]  {(b)};
        \node[left of= img2, node distance=0cm, xshift = -2.1cm, yshift=0, rotate = 90,font=\color{black}]  {Avg. $\|u(t) - u_\text{LQR}(t)\|^2$};
    \end{tikzpicture}
    \vspace*{-0.7cm}
    \caption{Comparison of average state and control error for LQR trajectory generation for the discretized double integrator with white Gaussian noise. The vanilla diffusion model, lacking dynamics awareness, exhibits high state and control errors. In contrast, our methods with known (Algorithm~\ref{algo:denoising-known-dynamics}) and unknown (Algorithm~\ref{algo:denoising-unknown-dynamics}) dynamics perform significantly better over the entire control horizon, demonstrating the effectiveness of incorporating dynamics into the diffusion process. Average over 100 test cases are shown.} \label{fig:LQR_diffusion}
\end{figure}
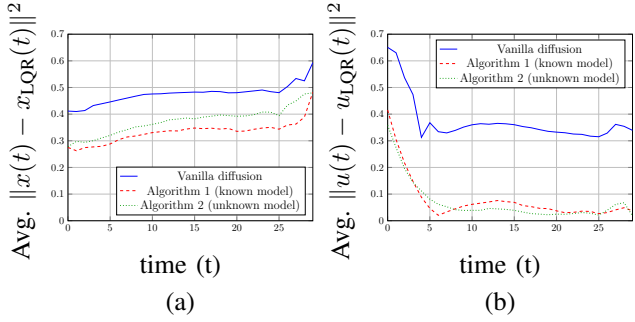

In Figure~\ref{fig:LQR_diffusion}, we compare the performance of three algorithms: vanilla diffusion (no dynamics awareness), our method with known dynamics (Algorithm~\ref{algo:denoising-known-dynamics}), and our method with unknown dynamics (Algorithm~\ref{algo:denoising-unknown-dynamics}). We plot the average state and control error compared to the true LQR controller for 100 test scenarios, where new $x_\text{init}$ and $x_\text{target}$ are sampled from $\mathcal{U}[-1,1]^4$ and $\mathcal{U}[-4,4]^4$, respectively. For each algorithm, 10 samples are generated per test case. The state error is defined as $\|x(t) - x_\text{LQR}(t)\|^2$, where $x_\text{LQR}(t)$ is the state generated by the LQR controller. The control error is defined as $\|u(t) - u_\text{LQR}(t)\|^2$, where $u_\text{LQR}(t)$ is the control input generated by the LQR controller. We observe that our proposed methods, algorithms~\ref{algo:denoising-known-dynamics} and~\ref{algo:denoising-unknown-dynamics} both outperform the vanilla diffusion model in terms of both state and control error. As the noise $w(t)$ non-zero, the performance of the diffusion model with dynamics awareness is slightly better than the one with unknown dynamics. This is because the diffusion model with dynamics awareness has access to the system dynamics, which allows it to generate more accurate trajectories. However, both algorithms perform significantly better than the vanilla diffusion model, which does not take into account the system dynamics.
\subsection{Waypoint tracking and collision avoidance}
\begin{figure*}[tbh]
    \centering
    \begin{tikzpicture}
        \node (img1) at (-9.4,0) {\includegraphics[width=0.180\textwidth]{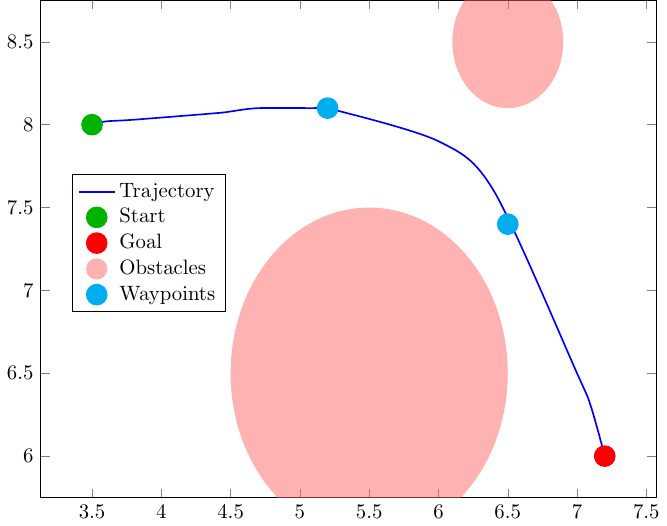}};
        \node[below of= img1, node distance=0cm, xshift = 0.0cm, yshift=-2.3cm,font=\color{black}]  {(a)};
        \node[below of= img1, node distance=0cm, xshift = 0.0cm, yshift=-1.8cm,font=\color{black}]  {x position};
        \node[left of= img1, node distance=0cm, xshift = -1.9cm, yshift=0cm,font=\color{black}, rotate  =90]  {y position};
        \node (img2) at (-5.8,0) {\includegraphics[width=0.1730\textwidth]{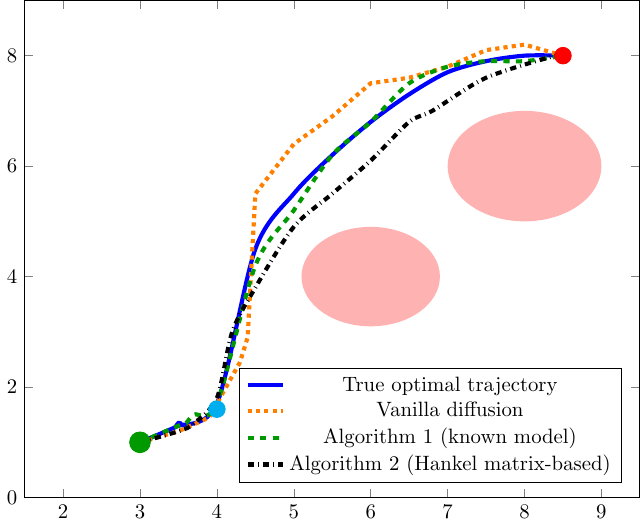}};
        \node[below of= img2, node distance=0cm, xshift = 0.0cm, yshift=-2.3cm,font=\color{black}]  {(b)};
        \node[below of= img2, node distance=0cm, xshift = 0.0cm, yshift=-1.8cm,font=\color{black}]  {x position};
        \node (img3) at (-1.25,0) {\includegraphics[width=0.250\textwidth]{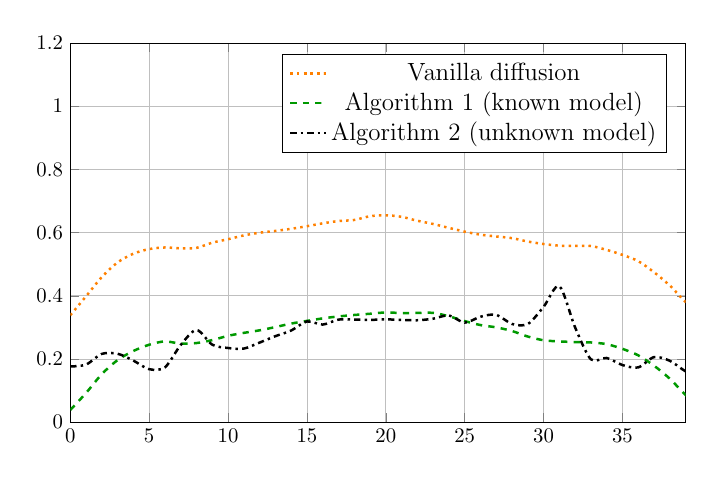}};
        \node[below of= img3, node distance=0cm, xshift = 0.0cm, yshift=-2.3cm,font=\color{black}]  {(c)};
        \node[below of= img3, node distance=0cm, xshift = 0.0cm, yshift=-1.8cm,font=\color{black}]  {time (t)};
        \node[left of= img3, node distance=0cm, xshift = -2.5cm, yshift=0, rotate = 90,font=\color{black}]  {Avg. $\|x(t) - x_\text{LQR}(t)\|^2$};
        \node (img4) at (3.8,0) {\includegraphics[width=0.25\textwidth]{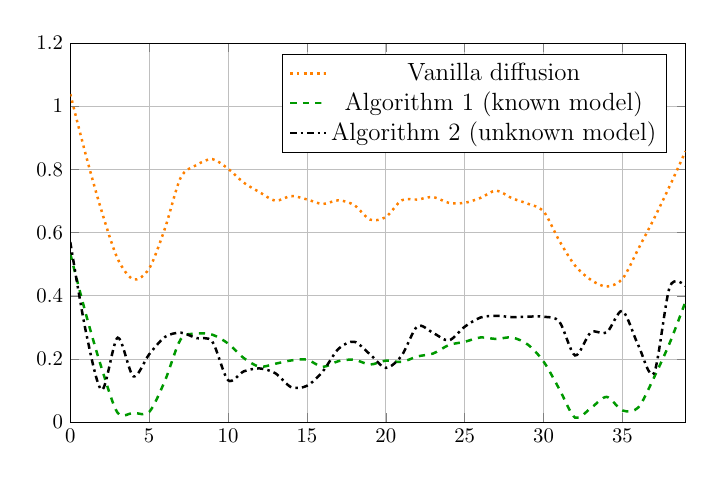}};
        \node[below of= img4, node distance=0cm, xshift = 0.0cm, yshift=-1.8cm,font=\color{black}]  {time (t)};
        \node[below of= img4, node distance=0cm, xshift = 0.0cm, yshift=-2.3cm,font=\color{black}]  {(d)};
        \node[left of= img4, node distance=0cm, xshift = -2.5cm, yshift=0, rotate = 90,font=\color{black}]  {Avg. $\|u(t) - u_\text{LQR}(t)\|^2$};        
    \end{tikzpicture}
    \vspace*{-0.65cm}
    \caption{Non-convex optimal control with waypoint tracking and obstacle avoidance. (a) Example of a dataset sample illustrating a solution to the non-convex optimal control problem~\eqref{eqn:complex_task}. The solid blue line depicts the numerically computed optimal state trajectory, blue dots represent the waypoints, and pale red circles indicate obstacles. (b) Comparison of sampled state trajectories generated by different algorithms. Algorithms 1 (known dynamics) and 2 (unknown dynamics) produce smooth, dynamics-aware trajectories closely resembling the true optimal trajectory, while the vanilla diffusion model deviates significantly. (c) Average state error versus time, and (d) average control error versus time, computed over 100 test cases. Algorithms~\ref{algo:denoising-known-dynamics} and~\ref{algo:denoising-unknown-dynamics} exhibit significantly lower errors due to their incorporation of system dynamics, demonstrating the effectiveness of our proposed framework.} \label{fig:complex_task}
\end{figure*}
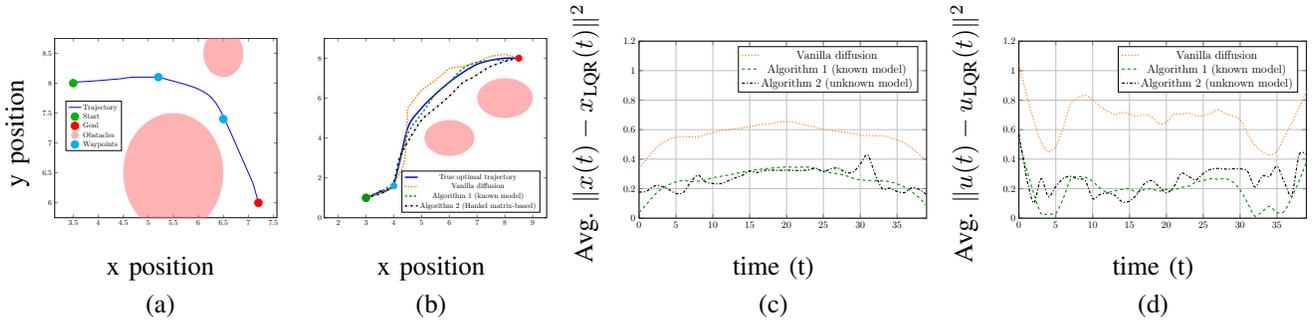
In this experiment, we seek to generate trajectories that arise as the solution to a non-convex optimal control problem for waypoint tracking and collision avoidance. The system is similar to the LQR problem in the previous experiment~\ref{sec:lqr}, but without noise $w(t) = 0$. In each experiment, we specify $V$ waypoints at positions $v_i \in \real^2$, and time instance $t_i$ to reach the waypoint. We include the target state $x_\text{target}$ as an additional waypoint with $t_i=T$. We also specify $O$ circular obstacles at positions $o_j \in \real^2$, with a radius $r_j \in \real_{\geq 0}$. The goal is to generate a trajectory that passes through the waypoints at times $t_i$ while avoiding the obstacles. Thus, the complex environmental variable $\E$ encodes information about the number and positions of waypoints, time to reach the waypoints, the number and positions of obstacles and their radii, and the target state $x_\text{init}$. The control problem is
\begin{align}
    \begin{split}
    \mathcal{R}(x(0:T),u(0:T),\E) &= \sum_{i=1}^{V+1} \| x(t_i) - v_i\|^2\\ 
    + \sum_{t=0}^{T-1}\Big( \|u(t)\|^2_R- \sum_{j=1}^O &\|(\|x(t) - o_i\|) - r_i\|^2\Big),\\
        \min_{u(t)} \quad  \mathcal{R}(x(0:&T),u(0:T),\E)\\
        \text{s.t. } \quad  x(t+1) &= Ax(t) + Bu(t),\\
         \quad \hspace*{0.55cm} x(0) &= x_\text{init}
    \end{split}\label{eqn:complex_task}
\end{align}
This is a non-convex optimal control problem that is typically handled in an MPC fashion. We synthetically generate 10,000 different environmental conditions $\E$, and solve the optimal control problem numerically to generate our training dataset. We also generate 100 different test data samples for evaluation. We use a similar neural network architecture as in experiment~\ref{sec:lqr} for each of the three different algorithms. The diffusion model is trained for 30,000 epochs with a batch size of 64 with a linear noising schedule. 

In Figure~\ref{fig:complex_task}(a), we represent a sample from the dataset. The blue line is the optimal trajectory of the system, and the blue dots denote the waypoints. The pale red circles are the obstacles. In Figure~\ref{fig:complex_task}(b), we compare the sampled trajectories for different algorithms. We observe that our proposed methods, algorithms 1 and 2 both produce smooth trajectories, whereas the vanilla diffusion model deviates significantly from the true optimal state trajectory. We generate 10 samples for each of the 100 test samples and compare the average state and control error with respect to the true optimal trajectory. The average state and control error for the three algorithms is shown in Figures~\ref{fig:complex_task}(c) and (d). We observe that our proposed methods, algorithms 1 and 2 both outperform the vanilla diffusion model in terms of both state and control error. Error drops at $t=5$ and $t=33$ correspond to waypoint tracking. As the noise $w(t)$ is zero, the performance of the diffusion model with dynamics awareness is slightly better than the one with unknown dynamics. This is because the diffusion model with dynamics awareness has access to the system dynamics, which allows it to generate more accurate trajectories. However, both algorithms perform significantly better than the vanilla diffusion model, which does not take into account the system dynamics. 

\section{Conclusion} 
This paper introduced a novel diffusion-based framework for generating dynamically admissible trajectories for linear systems, addressing both known and unknown dynamics scenarios. Our approach leverages a sequential prediction and projection mechanism, integrated seamlessly into the diffusion model's denoising process, to ensure that generated samples adhere to system constraints. We demonstrated the efficacy of our method in generating admissible trajectories that effectively solve complex control and planning problems, specifically showcasing its performance on a linear quadratic regulator (LQR) task and a challenging waypoint tracking and collision avoidance problem. The results highlight the significant improvement in trajectory accuracy and adherence to system dynamics compared to vanilla diffusion models. Future research directions include extending our framework to nonlinear systems, exploring efficient projection mechanisms for these systems, and investigating methods to accelerate the diffusion model's sampling process to enable real-time control applications.

\bibliography{refs}
\bibliographystyle{unsrt}



\end{document}